# Message du troisième type : irruption d'un tiers dans un dialogue en ligne

Ludovic Tanguy[1], Céline Poudat[2], Lydia-Mai Ho-Dac[1]

[1]CLLE : CNRS & Université de Toulouse –{ludovic.tanguy, lydia-mai.ho-dac}@univ-tlse2.fr

[2]Université Côte d'Azur, CNRS, BCL – celine.poudat@univ-cotedazur.fr

## Abstract

Our study focuses on Wikipedia talk pages, from a global perspective analyzing contributors' behaviors in online interactions. Using a corpus comprising all Wikipedia talk pages in French, totaling more than 300,000 discussion threads, we examine how discussions with more than two participants (multiparty conversation) unfold and we specifically investigate the role of a third participant's intervention when two Wikipedians have already initiated an exchange. In this regard, we concentrate on the sequential structure of these interactions in terms of articulation among different participants and aim to specify this third message by exploring its lexical particularities, while also proposing an initial typology of the third participant's message role and how it aligns with preceding messages.

## Résumé

Notre étude se concentre sur les pages de discussion Wikipédia, dans une perspective globale d'analyse des comportements des contributeurs dans leurs interactions en ligne. A partir d'un corpus comprenant l'ensemble des pages de discussion de Wikipédia en français, totalisant plus de 300 000 fils de discussion, nous nous intéressons à la façon dont une discussion à plus de deux participants (multilogue ou multiparty conversation) se met en place, en examinant le rôle de l'intervention d'un troisième protagoniste lorsque deux Wikipédiens ont déjà démarré un échange. Dans cette perspective, nous nous concentrons sur la structure séquentielle de ces interactions en termes d'articulation entre les différents intervenants, et visons à spécifier ce troisième message, en explorant ses particularités lexicales d'une part, et en proposant une première typologie du rôle du message du troisième intervenant et de la façon dont il s'articule avec les messages précédents.

**Mots clés :** Wikipédia, interactions, multilogues

## 1. Introduction

La présente étude a deux objectifs principaux : elle vise d'une part à explorer les particularités des interactions qui s'écrivent sur les pages de discussion Wikipédia, et d'autre part à apporter des éléments de description et de réflexion sur la manière dont s'engage une conversation en ligne à plusieurs (multilogue ou multiparty conversation). Dans cette optique, c'est l'arrivée et le rôle du troisième participant qui nous intéresse spécifiquement en ce sens qu'avec son intervention, on passe formellement du dialogue au multilogue. Si le dialogue est une forme privilégiée et cognitivement plus confortable que le monologue par exemple (Garrod and Pickering 2004), on sait que les multilogues sont des formes bien plus élaborées, complexifiant les rôles des participants qu'on a longtemps décrits comme speaker et addressee dans l'analyse des dialogues (Branigan 2006).

Ce qui nous intéresse dans la présente étude, c'est d'explorer ces aspects en lien avec les particularités des discussions Wikipédia qui influent sur la forme et les modes de déploiement des messages et des fils conversationnels. Ainsi, la visée de ces discussions, qui reste axée sur la réalisation d'une tâche commune, à savoir d'améliorer l'article encyclopédique associé, joue un rôle crucial dans la forme et les thèmes des échanges. Combiné au fait que ces discussions sont publiques, et donc accessibles et lisibles par tous, et qu'elles sont asynchrones, rédigées dans un format Wiki qui est unique dans le paysage de la communication médiée par les réseaux, le dispositif de communication particulier de Wikipédia régule sans aucun doute les interactions entre contributeurs, et le passage d'un dialogue à un multilogue. Enfin, la nature et le sujet des discussions, ainsi que les caractéristiques sociales des contributeurs impliquent un très bon niveau de rédaction et une modération dans les propos qui rendent le traitement et l'analyse plus aisés que pour d'autres types de réseaux sociaux. Si les situations de dialogue restent les plus classiques, avec des paires question/réponse, suggestion/avis, etc., le nombre de protagonistes dans une discussion est très variable, allant d'un seul utilisateur (qui n'obtient aucune réaction à son ou ses messages) à plusieurs dizaines lors notamment de consultations et de votes à l'échelle d'une communauté. Nous nous concentrons sur la structure séquentielle de ces interactions, en termes d'articulation entre les différents intervenants.

Après un premier aperçu à grande échelle des caractéristiques structurelles des fils de discussion avec un focus sur les conditions d'arrivée du troisième contributeur (2.), nous procédons à un examen plus détaillé du contenu des messages (3.) et présentons une première typologie du rôle du message du troisième intervenant et de la façon dont il s'articule avec les messages précédents que nous projetons sur un échantillon avec une annotation manuelle (4.).

## 2. Corpus et vue quantitative des données

Notre travail se fonde sur un corpus de discussions élaboré à partir du *dump* de la Wikipédia française téléchargé en septembre 2019. Nous avons sélectionné l'ensemble des pages de discussion associées aux articles, ainsi que leurs archives. Chaque page a été segmentée en fils de discussion selon les sections délimitées par les contributeurs et chaque fil en messages sur la base des signatures, indentations et autres marques de segmentation (Auteur, 2024). Seules les discussions comprenant au moins un message et deux mots ont été conservées. À chaque message est associé l'identifiant du contributeur (ou l'adresse IP des contributeurs anonymes), sa date d'écriture et bien entendu le contenu textuel. Les données sont encodées au format XML selon la norme TEI dédiée aux communications médiées par les réseaux (Beißwenger and Lüngen, 2020). Pour la présente étude, nous avons également éliminé toutes les discussions faisant intervenir un robot, en nous basant sur l'identifiant du contributeur. Au final, nous disposons d'un corpus de 302 475 discussions exploitables pour un total de 769 880 messages, écrits par 37 854 auteurs identifiés différents auxquels s'ajoutent 3143 IP différentes pour les utilisateurs anonymes, qui restent assez minoritaires.

L'exemple (1) fournit un aperçu, tel que proposé sur le site Wikipédia[1], d'un extrait du fil de discussion intitulé "Youtube" présent sur une version archivée de la page de discussion associée à l'article "Troubles au Tibet en mars 2008". Dans cet échange, un premier contributeur A annonce avoir effectué une modification de l'article qui est fortement désapprouvée par un contributeur B qui poste, quelques minutes après, un message plutôt agressif envers A qui répond dans la foulée (2 minutes plus tard). Un nouveau contributeur C (sous le pseudo Kyro) intervient avec le quatrième message pour exprimer son soutien à A et compléter la discussion avec de nouvelles informations que B semble totalement ignorer.

---

[1] https://fr.wikipedia.org/wiki/Discussion:Troubles_au_Tibet_en_mars_2008/archives_2009#Youtube

L'échange semble prendre fin au bout de 17 messages et 2 jours, par un message d'un contributeur D (pseudo Elnon) qui propose un consensus susceptible de pacifier le débat.

| | | |
|---|---|---|
| | | **Discussion:Troubles au Tibet en mars 2008/archives 2009** |
| | | Article  Discussion                    Lire  Modifier le code  Ajouter un sujet  Voir l'historique  Outils |
| | | < Discussion:Troubles au Tibet en mars 2008 |
| | | Youtube  [ modifier le code ] |
| 1 | A | J'ai supprimé les liens Youtube, après avoir lu la discussion suivante :[1]. Cela fait toujours autorité. --Rédacteur Tibet (d) 21 février 2010 à 19:26 (CET) [ répondre ] |
| 2 | B | Tu retire ces vidéos parce qu'elles montrent des témoignages accablants qui contredisent tes convictions. Le contenu de ces vidéos est une compilations d'extrait de témoignages diffusés sur de nombreuses télés publiques. La suppression est totalement arbitraire et abusive. Tu profite de la déstabilisation apportée par deux contributeurs pour relancer une guerre d'édition. --Aacitoyen (d) 21 février 2010 à 19:47 (CET) [ répondre ] |
| 3 | A | J'ai donné l'explication ci-dessus, merci de ne pas faire de procès d'intention.--Rédacteur Tibet (d) 21 février 2010 à 19:49 (CET) [ répondre ] |
| 4 | C | Ces videos n'ont pas été importer par les propriétaires des droits d'auteurs. Elles y sont donc illégalement. Les vidéos sont donc à retirer. Par contre, la vidéo peut être citée comme source … Cdlt, Kyro^(Tok Wiz Mi) le 21 février 2010 à 20:28 (CET) [ répondre ] |
| 5 | B | Si ce n'est pas tes convictions qui ont fait retirer ces deux vidéos, pourquoi n'as tu pas retiré les vidéos qui vont dans le sens de tes convictions dans le même chapitre, et qui posent la même question de copyvio ? --Aacitoyen (d) 21 février 2010 à 20:37 (CET) [ répondre ] |
| 6 | D | Question à Kyro : Je ne connais pas très bien la question, mais en quoi un simple lien actif (donc du code et non la vidéo elle-même) vers une vidéo située … --Elnon (d) 21… |
| … | | *10 messages et deux jours plus tard* |
| 17 | D | J'ai remis les intitulés des vidéos retirées, mais sans lien actif, de façon à rétablir l'équilibre des points de vue et la neutralité dans cette section.--Elnon (d) 23 février 2010 à 23:25 (CET) [ répondre ] |

*(1) Fil de discussion (titre : "Youtube") sur une archive de page de discussion liée à l'article "Troubles au Tibet…" impliquant un troisième intervenant ("Kyro"). La colonne 1 indique le numéro du message, la colonne 2 indique la lettre représentant le locuteur.*

Si un fil de discussion a par nature une structure arborescente, où chaque nouveau message s'insère en réponse à un message précédent comme pour les messages 2,3,4 et 6 en (1), le format Wiki utilisé dans les pages de discussion permet une grande liberté aux contributeurs qui peuvent insérer leur message à n'importe quel endroit (y compris à l'intérieur d'un message précédent ou avant celui-ci). Si l'on ajoute à cela le manque de fiabilité des marques de structures (e.g. indentations comme pour les messages 5 et 17 en (1)), la représentation fidèle du déroulement d'une discussion n'est pas toujours évidente (Poudat et al. 2017). Pour étudier plus facilement le déroulement de ces fils selon notre objet d'étude, nous avons dû "aplatir" les fils en considérant l'ordre linéaire de la discussion telle qu'elle apparaît sur la page. Cette approximation respecte l'ordre chronologique dans plus de 97% des paires de messages consécutifs.

De fait, nous considérons chaque fil de discussion comme une séquence de messages, chaque message ayant un rang (i.e. un ordre dans la linéarité de la discussion) et représentons les différents intervenants par une lettre indiquant leur ordre d'arrivée dans la discussion. Ainsi le début de l'échange en (1) suit un schéma de type "ABAC" correspond à la configuration où un troisième contributeur (C) intervient au rang 4, c'est-à-dire pour un quatrième message.

### 2.1. Vue d'ensemble des fils de discussion du corpus

Nous commençons par un aperçu quantitatif des 302 475 fils de discussions de notre corpus :
- 161 986 soit 53,5% ne contiennent qu'un seul message

- 174 009 soit 57,4% sont des discussions avec un seul utilisateur (parmi lesquels 12 023 soit 7% contiennent plusieurs messages)
- 83 196 soit 27,5% sont des discussions entre seulement deux utilisateurs
- 45 685 soit 15,1% impliquent plus de deux utilisateurs (jusqu'à un maximum de 43)

Ce sont ces derniers cas que nous examinons dans cet article, en nous concentrant plus précisément sur la place qu'occupe C dans la discussion.

### *2.2. Arrivée du troisième utilisateur*

Si l'on considère les fils de discussion qui contiennent au moins 3 messages, l'apparition d'un troisième protagoniste se produit dans 57,2% des cas (les autres cas étant des monologues ou des dialogues entre A et B). Autrement dit, les échanges faisant intervenir 3 utilisateurs sont très courants, et même majoritaires lorsqu'une discussion s'étend sur plus de deux messages.

Lorsqu'un troisième utilisateur intervient, il survient dès le troisième message dans 63% des cas. En d'autres termes, 63% des échanges impliquant un 3e utilisateur suivent un schéma commençant par ABC. Dans certains cas, la discussion peut se produire après un long dialogue entre A et B (dans le cas le plus extrême observé, C arrive au rang 43).

Si l'on regarde les profils temporels[2], l'intervention de C se produit après un délai médian de 1,25 jours après le premier message, et un délai médian de 7,0 heures après le message précédent son arrivée, quel qu'en soit l'auteur (A ou B) et le rang (2 ou plus). Ce délai est significativement supérieur (test de Mann-Whitney, $p<0.01$) à celui que nous avons mesuré entre deux messages consécutifs sur l'ensemble du corpus (médiane de 3,2 heures) et encore plus si l'on ne considère que les messages d'un rang supérieur ou égal à 3 (2,3 heures). Autrement dit, l'intervention d'un tiers dans un dialogue installé se fait avec un temps de réaction (ou de réflexion) plus long, que l'on peut a priori expliquer en partie par le temps nécessaire pour prendre connaissances des échanges précédents et des éventuelles différences entre les points de vue exprimés.

Si l'on examine maintenant la façon dont la discussion se poursuit *après* l'arrivée de C, nous avons observé les tendances suivantes. Dans 30,8% des cas, le message de C est le dernier de la discussion[3]. Par comparaison, une discussion n'impliquant qu'un dialogue entre A et B se termine dans 34,2% des cas après 3 messages. Autrement dit, l'arrivée de C est susceptible de prolonger la discussion au-delà de son message, comme ce que l'on observe en (1).

La figure 1 propose une vision d'ensemble des différentes interventions des protagonistes dans une discussion Wikipédia, de son début (premier message de A, à gauche) jusqu'à son achèvement au moment de la capture des données (nœuds rectangulaires). Les nœuds en gris représentent l'ensemble des données considérées dans la suite de l'étude.

---

[2] Nous utilisons systématiquement la médiane pour mesurer les tendances centrales des durées séparant les messages, étant donné la présence importante de valeurs extrêmes qui biaisent la moyenne. Par exemple, nous avons pu identifier des discussions s'étalant sur plus de 15 ans, avec des pauses de très longue durée.

[3] Cela ne signifie pas que l'échange est terminé, étant donné qu'une discussion en ligne n'a pas de fin formelle et qu'une réaction peut arriver des années plus tard.

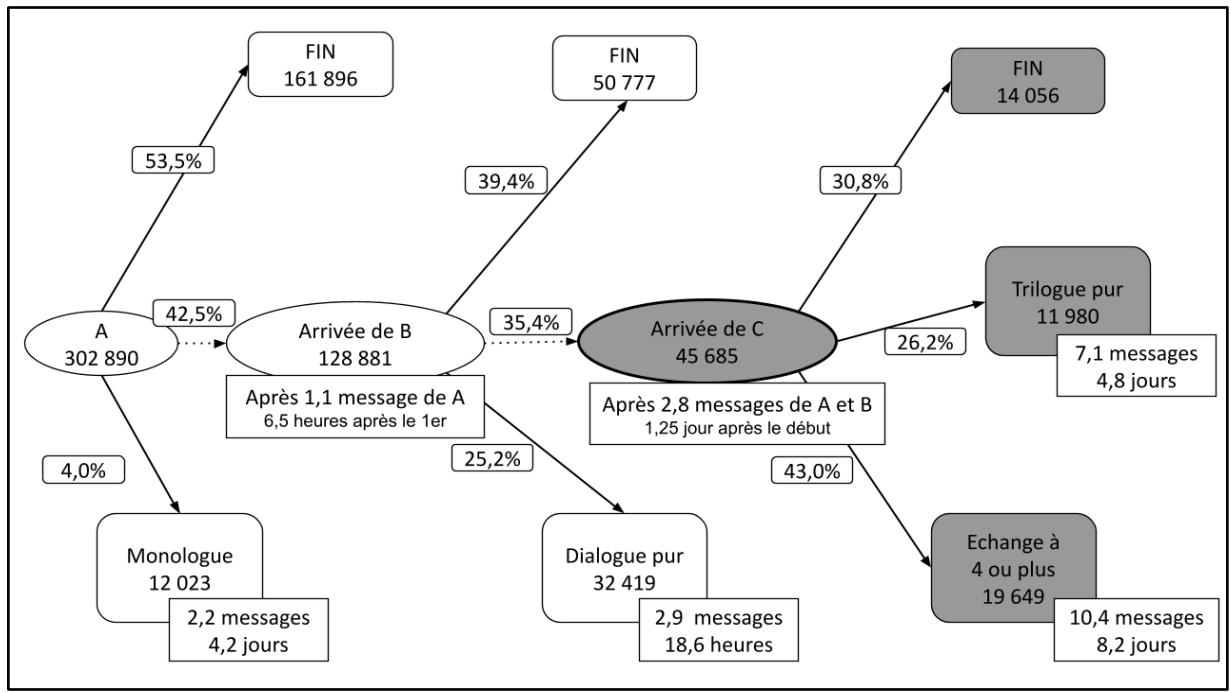

Figure 1: *Vue d'ensemble des différents types de fils de discussion en fonction des protagonistes. Les nombres sans unité indiquent les effectifs des fils dans le corpus ; les nombres de messages sont des moyennes et les durées des médianes. Les cases grisées indiquent les données de l'article.*

## 3. Exploration du contenu des premiers messages de A, B et C

Il nous a d'abord semblé pertinent de dégager les particularités lexicales du premier message de C. Dans cette perspective, nous avons extrait l'ensemble de ces premiers messages que nous avons comparé aux premiers messages postés par A et B sur l'ensemble des 45685 fils. Un calcul des spécificités (mots formes) de chaque ensemble réalisé par TXM[4] et les premières spécificités positives et négatives sont restituées dans le tableau 1.

Les premiers messages de A, B et C se singularisent d'abord de manière étonnamment nette, ce qui nous conforte dans la pertinence de notre approche. Le premier message de A, qui ouvre le fil et qui est donc plus long que les messages de B et C, se distingue par l'usage de termes phatiques, marquant l'attente d'un retour de la part des autres contributeurs relativement à une proposition de départ (*pensez, Quelqu'un, -il*, que l'on relève dans des séquences de type *faut-il, est-il, Est, -on*). On observe que les marqueurs d'accord et de désaccord le spécifient négativement, ce qui montre que ce premier message se voudrait peut-être neutre, ou du moins ouvert. Le premier message de B, qui répond donc à A, contient pour sa part une proportion significative de marqueurs de deuxième personne, avec de manière intéressante des marqueurs d'accord plutôt que de désaccord (premiers cooccurrents du mot-forme *accord*: *fait* +241 pour *tout à fait d'accord*, *je, suis, Entièrement* (d'accord) +127), ce qui n'est pas surprenant, l'accord étant évidemment la forme préférée. Le premier message de C, qui nous intéresse spécifiquement, s'inscrirait également d'abord dans une dynamique d'alignement, mais de manière peut-être moins nette car soit mitigée (*contre* + 63,4, qui peut d'ailleurs advenir dans des formes plus ou moins directes comme *je ne suis pas contre*, est plus spécifique que *Pour* +47,7 et on relève la présence significative de la négation *n'*), soit possiblement plus argumentée (*effectivement*) - il faut garder à l'esprit que C arrive dans une situation complexe puisque deux contributeurs ont démarré une discussion, qui peut d'ailleurs être installée depuis de nombreux tours de parole lorsqu'il intervient (cf section 2).

---

[4] TXM (https://txm.gitpages.huma-num.fr/textometrie/index.html) mobilise un calcul fondé sur la distribution hypergéométrique, voir Poudat et Landragin 2019: 170-171.

|  | A | Indice | B | Indice | C | Indice |
|---|---|---|---|---|---|---|
| **Spécificités positives** | depuis | 1000,0 | tu | 1000,0 | ☐☐[5] | 118,7 |
|  | révisions | 248,8 | vous | 154,1 | c' | 77,5 |
|  | propose | 243,8 | pas | 121,5 | contre | 63,4 |
|  | -il | 231,0 | toi | 95,6 | Pour | 47,7 |
|  | Wikipédia | 212,7 | te | 94,5 | pas | 45,2 |
|  | pensez | 144,1 | accord | 91,6 | effectivement | 43,5 |
|  | Bonjour | 116,5 | Oui | 87,8 | aussi | 42,5 |
|  | Quelqu'un | 90,9 | Tu | 85,7 | est | 39,5 |
|  | Qu' | 73,5 | c' | 85,4 | pour | 38,5 |
|  | Est | 71,4 | as | 81,2 | rejoins | 37,8 |
|  | -on | 64,2 | est | 81,0 | n' | 28,8 |
| **Spécificités négatives** | tu | -1000,0 | propose | -84,1 | -il | -82,9 |
|  | accord | -1000,0 | -il | -66,0 | Bonjour | -77,9 |
|  | c' | -294,6 | signé | -51,2 | propose | -77,7 |
|  | vous | -281,5 | pensez | -50,2 | pensez | -48,2 |
|  | pas | -275,5 | révisions | -42,7 | article | -42,1 |
|  | Oui | -207,1 | la | -41,7 | Quelqu'un | -33,9 |
|  | est | -202,0 | l' | -36,2 | suggérée | -32,0 |
|  | Pour | -182,6 | Quelqu'un | -28,0 | wikipédia | -31,9 |
|  | effectivement | -178,1 | Est | -26,4 | Qu' | -31,1 |
|  | contre | -163,5 | -on | -24,7 | transfert | -28,9 |

*Tableau 1 : Spécificités des mots-formes des premiers messages de A, B et C*

## 4. Objectif du troisième type

Afin de décrire plus précisément les interventions de C, nous avons retenu trois critères qui nous semblent pertinents: le(s) destinataire(s) du message, l'alignement éventuel de l'intervenant avec la position défendue par les intervenants précédents A et/ou B et l'objectif principal de l'intervention. La section suivante décrit et illustre les catégories distinguées pour chacun de ces trois aspects, qui ont été soumis à une tâche d'annotation exploratoire dont nous présentons les résultats majeurs.

### 4.1. Grille d'annotation

L'annotation des destinataires distingue deux cas de figure selon que C s'intègre de façon **générale** dans la discussion, comme avec la première intervention de C en (1) ; ou que C réagisse de façon **ciblée** à un message précédent comme avec le sixième message de (1) : *"Question à [C] : [...]"*. Globalement, on considère une intervention de C comme ciblée si elle peut être interprétée indépendamment du reste de la discussion.

L'annotation de l'alignement consiste à décrire comment la première intervention de C se positionne par rapport aux positions exprimées par A et B. Nous avons regroupé les situations en quatre cas : l'**harmonie** (A et B sont d'accord et C va également dans leur sens), la **prise de parti** (A et B s'opposent et C vient soutenir l'un des deux), l'**opposition** (C prend une nouvelle position qui s'oppose à A et à B, que ceux-ci soient d'accord entre eux ou non) et la **neutralité** (qui recouvre également les discussions où C n'a aucune raison de se positionner, la discussion pouvant être un simple espace de coordination par exemple).

---
[5] La coche indique un accord ou une action réalisée.

Concernant l'annotation de l'objectif de l'intervention de C, nous avons pu distinguer 10 catégories qui permettent d'éclairer son comportement :

1. **vote** : les discussions Wikipédia peuvent être le lieu de votes destinés à valider ou invalider une proposition. De façon générale, ces votes s'inscrivent dans un temps court et impliquent de nombreux utilisateurs qui ne s'expriment qu'une fois et de manière généralement lapidaire.
2. **rapport d'activité** : C rapporte une action qu'il a réalisée (généralement suite aux échanges entre A et B), comme dans le message 1 et 17 de (1) : *"J'ai remis les intitulés des vidéos retirés…"*
3. **complément** : C ajoute des informations au sujet discuté au moment de son intervention sans prendre partie en cas de désaccord entre A et B.
4. **appui** : C soutient explicitement l'opinion ou la proposition exprimée par A ou B sans ajouter d'information ou d'argument. Ce soutien peut s'exprimer de manière indirecte par une proposition en faveur de l'opinion soutenue ou par un remerciement envers son auteur mais en faisant référence à cet intervenant.
5. **appui et complément** : il est fréquent que C complète son appui par une information comme dans (1) où C soutient A tout en apportant un complément d'information sur le statut et l'utilisation des liens Youtube en jeu dans le désaccord entre A et B.
6. **opposition** : à l'inverse des deux catégories précédentes, C s'oppose à l'opinion ou une proposition exprimée par A et/ou B. Cette opposition est systématiquement accompagnée d'une argumentation (au contraire du vote précédent pour lequel C peut simplement écrire "Contre") et parfois d'un complément d'information.

   *(2) Attention [A], vous m'avez bien mal lu. [B] n'est en rien dans mon interruption de discussion. Au contraire, je le soutiens. Pour être encore plus clair, c'est votre position si arrêtée et celle de [X] qui m'ont usées. [...]*

7. **question** : C intervient en demandant des compléments d'information, un avis ou une action de la part des autres participants.
8. **réponse** : C fournit une réponse factuelle (et non un avis) à une question posée.
9. **pacification** : C cherche à pacifier une situation de désaccord qui devient conflictuelle. Cette pacification peut prendre la forme d'une proposition de compromis, d'un appel à la trêve ou d'un rappel de la manière dont de telles situations sont gérées dans la communauté Wikipedia.
10. **ouverture** : C apporte une information ou une réflexion qui n'est pas directement liée au sujet discuté par A et/ou B mais qui propose une prise de distance, une mise en perspective, ou va ouvrir un débat parallèle à celui de la discussion initiale.

### 4.2. Annotation des messages du troisième type

Le message de C peut advenir à des positions et dans des situations très différentes, comme nous l'avons montré dans la Figure 1. Afin de couvrir un éventail large de situations, nous avons constitué deux échantillons différents : l'**échantillon 1** contient une sélection aléatoire de discussions avec au moins une intervention de C (parmi les 45 685 fils détectés). L'**échantillon 2** rassemble une sélection aléatoire de fils où C n'intervient qu'au bout du dixième message ou après, ce qui est le cas pour 626 fils. Alors que l'**échantillon 1** aura tendance à contenir des échanges standards dans lesquels C intervient plutôt en début de discussion (pour rappel, C intervient au rang 3 dans 63% des échanges impliquant C), l'**échantillon 2** permet d'observer les cas plus rares où C intervient après qu'un long dialogue se soit installé entre A et B. L'échantillon 2 ne contiendra évidemment pas de vote (qui implique que C intervient au troisième message) mais potentiellement des objectifs plus spécifiques étant donné la longueur de l'échange préalable entre A et B qui a notamment plus de chance d'être de nature conflictuelle, comme l'avaient montré Denis et al. (2012).

Une première annotation exploratoire a été réalisée par trois annotateurs sur 84 fils (42 de chacun des deux échantillons) sur les trois dimensions évoquées précédemment (cible, alignement et objectif). Le calcul d'un accord inter-annotateur ne nous a pas semblé adapté à ce stade, l'adjudication ayant suivi ayant également permis de fixer la grille d'annotation. Cette annotation reste difficile, surtout pour ce qui concerne l'alignement et les discussions longues (échantillon 2) où A et B discutent parfois de sujets difficiles à cerner (débat entre spécialistes, renvoi à des échanges ou des modifications d'articles précédents impliquant une communauté d'éditeurs, etc.). Nous pouvons néanmoins proposer une première vue quantifiée des différentes tendances.

Concernant les destinataires, l'intervention de C n'est ciblée que dans 33% (échantillon 1) et 27% (échantillon 2) des cas. Il s'agit donc d'une situation relativement minoritaire sans variation significative entre les deux échantillons : l'arrivée de C dans une discussion concerne majoritairement les deux protagonistes.

Concernant l'alignement, la figure 2 montre la répartition des fils annotés de chaque échantillon dans les 4 catégories considérées.

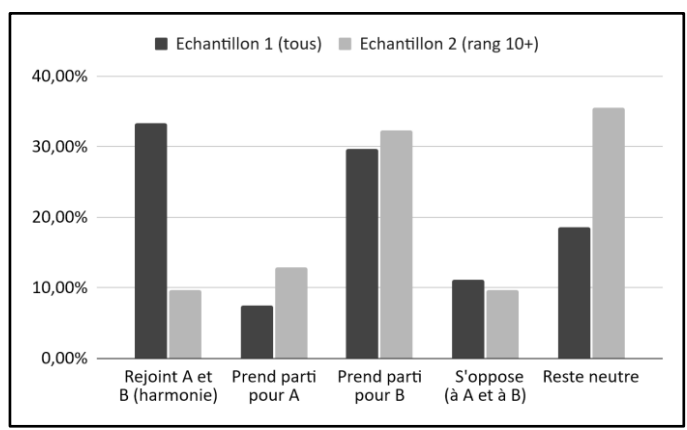

*Figure 2 : Alignement de C dans le fil, proportion dans chaque échantillon annoté*

Le cas majoritaire dans les deux échantillons est celui où C s'aligne avec A et/ou B : que ce soit dans une discussion où tout le monde est d'accord (33% d'harmonie dans l'échantillon 1, 10% seulement dans l'échantillon 2) ; ou dans une discussion qui oppose A et B (37 et 45% au total), avec dans ce cas une nette préférence pour s'aligner avec B (70% des cas de prise de parti, sans variations remarquables entre échantillons). Il semblerait que quand il faut choisir entre l'intervenant à l'origine de la discussion (A) et celui qui a réagi (B), C s'aligne donc plutôt avec le second. Une spécificité des interventions tardives semble aussi se dégager, avec moins d'accord global contrebalancé par plus de neutralité. Nous allons voir ces différences plus clairement en examinant les objectifs.

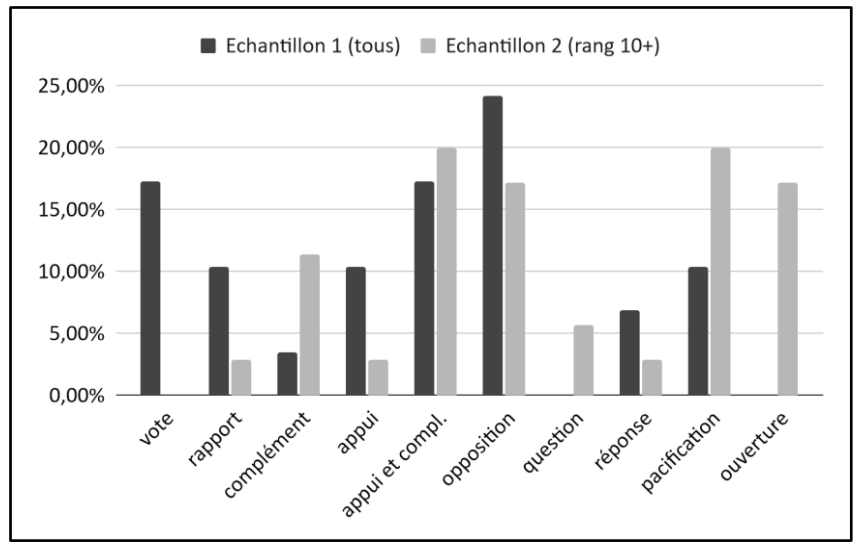

*Figure 3 : Objectif principal du premier message de C, proportion dans chaque échantillon annoté*

L'intervention de C exprime le plus souvent un soutien à A et/ou B: c'est la visée d'un quart des interventions de C les deux échantillons confondus si l'on fusionne les deux catégories "appui" et "appui et complète". A l'inverse, C intervient ensuite pour s'opposer (22% des cas). Que ce soit pour soutenir (et compléter) ou s'opposer, ces premiers résultats vont dans le sens des spécificités observées dans la section précédente, et éclairent la présence significative des termes *contre, Pour et rejoins*. Dans près de la moitié des fils (49%), C intervient pour prendre parti pour/contre A et/ou B avec une différence notable entre entre échantillons, comme le montre la Figure 3.

On observe en effet nettement moins d'"appui" simple et d'"opposition" dans l'échantillon 2 (3% et 17%) que dans l'échantillon 1 (13% et 29%). Il semble donc que lorsque C arrive au bout d'une discussion déjà bien établie (et potentiellement conflictuelle), un soutien non constructif ou une opposition frontale est plus délicate.

De manière générale, les différences obtenues entre les deux échantillons sont assez cohérentes: on ne relève bien entendu aucun vote dans l'échantillon 2. Au contraire, lorsqu'un dialogue long s'installe entre A et B, il devient plus compliqué pour C d'intervenir en répondant à une question posée. On a vu dans la section 3 que le premier message de A est souvent général, s'adressant volontiers à tous (*Qu'en pensez-vous?*) tandis que le message de B est déjà plus ciblé. Au bout d'une dizaine d'échanges entre A et B, on peut considérer que la forme *dialogue* l'a emporté. Malgré la nature publique de la discussion, il peut être compliqué de survenir dans un dialogue sans une bonne raison (proposer un complément d'information, une ouverture, pacifier la discussion).

## 5. Conclusion et perspectives

Nous avons exploré le comportement du troisième intervenant dans les fils de discussion Wikipédia en multipliant les angles d'observation et en adoptant une approche mixte, combinant analyse de données à grande échelle, étude des schémas d'interactions et annotation manuelle de typologies fines pour espérer à terme annoter de manière automatique certains phénomènes. Après avoir caractérisé les conditions situationnelles de l'arrivée de C dans une conversation (temps, configuration conversationnelle) par rapport à l'ensemble des discussions de Wikipédia, nous avons examiné le contenu de ses premiers messages par rapport aux premiers messages de A et B. La question de l'accord/désaccord a notamment semblé spécifique à C, qui endosse potentiellement un rôle d'arbitre dans le cas d'un désaccord entre A et B. Afin de caractériser finement le rôle de C en mettant en évidence des aspects que les deux analyses précédentes ne permettent pas d'observer directement, nous avons mis en œuvre une tâche d'annotation exploratoire des données qui nous a permis de proposer une typologie explicative des comportements de C, avec une première quantification.

Si nous travaillerons évidemment à affiner notre typologie et à l'ajuster en considérant d'autres échantillons correspondant à d'autres situations conversationnelles (*e.g.* distinction des cas de multilogues et de multi-dialogues, cas où le message de C clôture le fil, *etc.*), nous insisterons pour finir sur l'intérêt de notre approche, et sur son caractère inédit dans les champs de l'analyse des interactions et de l'étude des genres du Web : les conversations multilogues ont surtout été observées par l'analyse conversationnelle, qui se concentre plutôt sur les interactions orales, et sur des phénomènes d'alignement linguistique et de typologie des participants (e.g. ratifiés, *overhearer*, cf. Braningan 2006), dans des cadres micro-analytiques (voir Meredith *et al.* 2021 pour l'étude des interactions du Web). Notre perspective, centrée sur les interactions écrites, vise à étudier ces phénomènes à grande échelle, ce que permet aujourd'hui la communication médiée par les réseaux en produisant des masses de données. Nous continuerons ainsi à développer cette approche globale dans les années qui viennent en développant progressivement la dimension multilingue de notre approche, qui s'y prête ainsi particulièrement - un travail de comparaison avec l'anglais est actuellement en cours.

## Bibliographie